\documentclass[11pt]{article}

\usepackage[preprint]{acl}
\usepackage{float}
\usepackage{times}
\usepackage{latexsym}
\usepackage{card}
\usepackage[T1]{fontenc}

\usepackage[utf8]{inputenc}

\usepackage{microtype}

\usepackage{inconsolata}

\usepackage{graphicx}
 \usepackage{booktabs}

\usepackage{comment}
\usepackage{multirow, multicol}
\usepackage{subcaption}
\usepackage{amsmath}
\title{Linguistic Distance Segregates Latent Representations \\in Automatic Speech Recognition Systems}

\author{Ting-Hui Cheng, Line Katrine Harder Clemmensen\and Sneha Das \\
Department of Applied Mathematics and Computer Science \\
        Technical University of Denmark\\
        \texttt{\{tiche, lkhc, sned\}@dtu.dk}
        }

\begin{document}

\newpage
\maketitle
\begin{abstract}
  While automatic speech recognition (ASR) models have achieved remarkable improvements in recent years, performance disparities persist across different speaker populations. One such disparity is for speakers whose first languages (L1) are from families distant from English. 
  This paper investigates the relationship between first language background and English ASR performance. Through empirical analysis, we observe that the correlation between speakers' L1 distance and ASR error rates yields a systematic effect on English Speech, with its strength varying across datasets and models. 
 This association is statistically significant in a follow-up analysis accounting for dataset-level variation in Tweedie mixed-effects models ($p<0.001$ across evaluated models). In addition, analysis of the latent space reveals a L1-based spatial segregation across deeper acoustic layers in the majority of evaluated architectures\footnote{More details about data and code are available: \href{https://github.com/tinghui8576/LD-Bias-EngASR}{https://github.com/tinghui8576/LD-Bias-EngASR}}.  
\end{abstract}

\section{Introduction}
Automatic Speech Recognition (ASR) systems are widely deployed in real-world applications. Ensuring robust performance across diverse speaker populations is therefore essential. Yet, performance disparities across speaker subgroups continue to be reported, raising questions about when and who the systems underperform for?

In this work, we define L1 speakers as those whose first language is English, and L2 speakers as those who acquired English as a second language. Prior work has documented ASR performance disparities across various speaker attributes. Differences have been observed with respect to gender~\cite{
attanasio2024twists} and racial background~\cite{
zolnoori2024decoding}. Beyond these demographic factors, L2 speakers exhibit higher Word Error Rates (WER) than L1 across commercial English ASR systems (Google, Amazon, Microsoft)~\cite{dichristofano2022global} and open-source models such as Whisper~\cite{graham2024evaluating}. Similar disparities have also been reported for Swedish~\cite{cumbal2024you} and Korean~\cite{10848815}.

\begin{table}[t]
    \centering
    \caption{Overview of the datasets, including total samples (hours) and the number of accents.}
    \label{tab:dataset_properties}
    \resizebox{\columnwidth}{!}{  
    \begin{tabular}{lcc}
        \toprule
        \textbf{Dataset} & \textbf{N. Samples (hr)} & \textbf{Accents} \\
        \midrule
        \textbf{Speech Accent Archive (SAA)}~\cite{weinberger2015speech} & 2,015 (15.5) & 17 \\
        \textbf{Fair-Speech}~\cite{veliche2024towards}              & 26,471 (54.5) & 27 \\
        \textbf{L2-ARCTIC (L2Arc)}~\cite{zhao2018l2arctic}                 & 26,867 (27.1)    & 6  \\
        \textbf{EdAcc}~\cite{sanabria23edacc}                     & 17,965 (32.7)    & 20 \\
        \textbf{ALLSSTAR} \cite{bradlow2010allsstar}                & 699 (20.9)    & 22 \\ 
        \textbf{AFRISPEECH-200 (Afri200)}~\cite{olatunji2023afrispeech} & 6318 (18.77) & 	108 \\
        \bottomrule
    \end{tabular}
    }
\end{table}

These performance gaps are commonly attributed to speaker characteristics, including geographic origin~\cite{jahan2025unveiling} and first-language influence~\cite{chan2022training}. Prior research suggests that L1-L2 disparities may stem from systematic structural differences between languages. Linguistic research has shown that L1-L2 distance plays an important role in second language processing and acquisition~\cite{kuperman2025does}. Similar effects have been observed in NLP systems, where linguistic proximity impacts cross-lingual transfer performance, such as in automatic abusive language detection~\cite{eronen2022transfer}. These findings suggest that linguistic distance may systematically impact model generalization across languages. However, its role in explaining disparities across ASR remains limited. 

In this work, we examine whether L1-L2 ASR performance gaps on English speech are systematically associated with structural Linguistic Distance ($LD$) between speakers’ L1 and English. Furthermore, we conduct a comparative analysis across distinct ASR architectures to evaluate their respective sensitivities to L2 speech. We observe that as input representations propagate into deeper layers, all evaluated models maintain a clear segregation between L2 acoustic representations and those of English L1 speakers. This systematic spatial separation implies that current state-of-the-art architectures isolate accented features rather than robustly normalizing them, emphasizing the need for enhanced accent-invariant representation learning in future acoustic frameworks.

\section{Related Work}
Several studies have examined the impact of speakers’ L1 on ASR performance, showing that recognition accuracy varies with speaker language and accent~\cite{chan2022training, liu2025evaluating, jahan2025unveiling}. While prior work documents L1-L2 performance disparities, it has largely focused on quantifying these gaps. 
\citet{issa2026measuring} investigated Whisper large-v3’s performance on L2 Egyptian Arabic speech and revealed existing correlation across WER and high accentedness (rated by human) but comprehensibility only had a marginal positive association. \citet{bartelds2021measuring} showed that there exists a correlation between acoustic distances in neural speech representation and human native-likeness judgments. 
\citet{tsoukala2026extending} revealed a clear performance gradient with dialectal distance from Standard Modern Greek. Similar research has been shown in Finnish~\cite{11385259}. \citet{toro2025neighbors} further showed that language embeddings cluster meaningfully, with distances correlating with geographic and lexical distances across languages, while
\citet{prasad2020accents} found that accent-related information is largely encoded in early ASR layers. While prior studies link embeddings to linguistic and accent features, their relevance to L2-ASR performance disparities remains unexplored.

\begin{figure}[t]
    \centering
    \includegraphics[width=\linewidth]{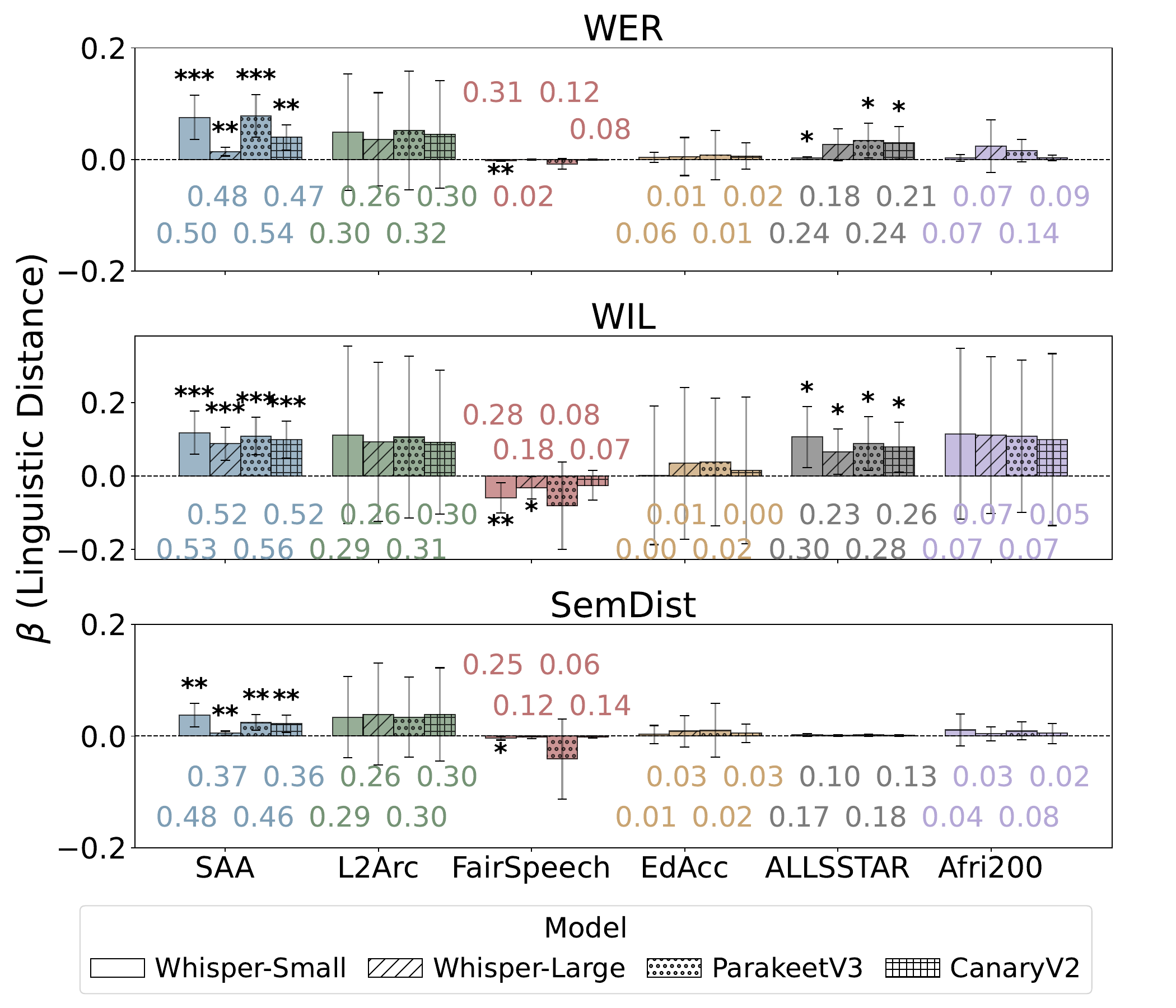}
    \caption{Coefficient ($\beta$) for Linguistic Distance ($LD$) relative to English ASR model performance in WER, WIL and SemDist metrics. Error bars represent 95\% confidence intervals, and the value shown is the adjusted $R^2$ (*: $p<0.05$, **: $p<0.01$, ***: $p<0.001$). Statistical assumptions were verified and met.}
    \label{fig:coef}
\end{figure}

\section{Methodology}
\paragraph{Language linguistic distances:}
We evaluate two speaker populations: first language speakers in English (L1) and speakers who acquired English as a second language (L2).
To quantify the linguistic dissimilarity between each speaker's first language and English, we apply method proposed by~\citet{vincent2020stochastic}\footnote{\url{http://www.elinguistics.net/Compare_Languages.aspx}}, which provides a single composite score of linguistic relatedness between languages for subsequent analysis. The resulting linguistic distance ($LD$) score ranges from 0 to 100 (identical to completely unrelated languages).

\paragraph{Regression model:}
We performed a regression analysis\footnote{From SciPy package} with $LD$ as the main parameter to quantify the impact of speakers' L1 transfer on English ASR performance.  
\begin{equation} \label{eq:regression}
\begin{split}
Y_i^{metric} &= \beta_{const} + \beta_{LD} x_{LD,i}+ \epsilon,\\
\end{split}
\end{equation}
where $Y_i^{metric}$ is the mean of performance metric evaluated on language $i$; $\beta_{const}$ represents constant intercept term, capturing the model's baseline error rate for L1 speakers; $x_{LD,i}$ denotes the $LD$ from language $i$ to English; $\beta_{LD}$ is the slope coefficient quantifying model's performance sensitivity to increasing linguistic distance; and $\epsilon_i \sim \mathcal{N}(0, \sigma^2)$ is the residual error term. Both $Y_i^{metric}$ and $x_{LD,i}$ are normalized via Min-Max scaling\footnote{From Scikit-learn package} to ensure a uniform baseline for comparison.

\paragraph{Generalized Linear Mixed Model:}

We conduct a follow-up investigation into the association between speakers' L1 $LD$ from English and ASR performance across datasets by accounting for dataset differences. Towards that, we employ a generalized linear mixed model with a Tweedie likelihood\footnote{From GBBoost package}~\cite{ma2007nested}, shown in Equation~\ref{glmm}. The Tweedie family is suitable for continuous, non-negative response variables and can accommodate a point mass at zero through its compound Poisson--Gamma representation, making it appropriate for modeling ASR error-related measures.
\begin{equation}
\label{glmm}
\begin{split}
Y_{gi}^{\mathrm{WER}} \mid u_g
&\sim \operatorname{Tweedie}(\mu_{gi}, \phi, p), \\
\log(\mu_{gi})
&= \beta_{0} + \beta_{1}x_{LD,i} + u_g, \\
u_g &\overset{\mathrm{i.i.d.}}{\sim}
\mathcal{N}(0, \sigma_u^2), 
\end{split}
\end{equation}
where $Y_{gi}^{\mathrm{WER}}$ denotes the WER for observation $i$ in
dataset $g$, $\mu_{gi}$ is its conditional mean with $Ng$ represents the number of observation within the group, $\sigma_u^2$ is the variance component, $\phi$ is the
dispersion parameter, and $p$ is the Tweedie power parameter constrained to $1 < p < 2$. The conditional mean and variance are given by
\[
 E\!\left[Y_{gi}^{\mathrm{WER}} \mid u_g\right] = \mu_{gi},
\quad
\operatorname{Var}\!\left(Y_{gi}^{\mathrm{WER}}\mid u_g\right)
= \phi\mu_{gi}^{p}.
\]
The coefficient $\beta_1$ represents the population-level $LD$-WER association, while $u_g$ captures
dataset-specific deviations from overall intercept. The model trends for WIL and SemDist are presented in Appendix~\ref{tweedie}.

\paragraph{Embedding distance and correlation:}
To evaluate how L2 phonetic variants are encoded dynamically within the internal representations across the network's depth, we compute the Spearman rank correlation coefficient~($\rho$) between $LD$ and the layer-specific~($l$) Embedding Distance~($\text{ED}_{\text{l}}$) of L2 individual utterances relative to a English L1 speakers' utterance baseline. Following similar approaches utilized in recent literature to measure embedding distance~\cite{durrheim2023using, bolukbasi2016man, ohamadike2025whose}
, we apply cosine distance as our core metric. For any given target utterance embedding vector $E_{\text{S},l}$, we calculate its cosine distance to the centroid of the L1 English reference embeddings, $C(E_{\text{E},l})$, as:
\begin{equation}
ED_{l} 
= 1 -
\frac{
C(E_{\text{E},l}) \cdot E_{\text{S},l}
}{
\|C(E_{\text{E},l})\| \, \|E_{\text{S},l}\|
}
\end{equation}
where the L1 English centroid $C(E_{\text{E},l})$ represents the geometric mean of all L1's utterance embeddings in all evaluating datasets within the corresponding network layer $l$, calculated as:
\begin{equation}
C(E_{\text{E},l}) = \frac{1}{N_{\text{E}}} \sum_{i=1}^{N_{\text{E}}} E_{\text{E},l, i}
\end{equation}
Here, $N_{\text{E}}$ denotes the total number of L1 speaker utterances, and $E_{\text{E}, l, i}$ represents the embedding vector of the $i$-th L1 speaker sample in layer $l$. The layer-wise Spearman rank correlation ($\rho_l$) between the embedding displacement distribution and the structural $LD$ is calculated as:
\begin{equation}
\rho_l = \frac{\text{cov}(R[LD], R[ED_{l}])}{\sigma_{R[LD]}\sigma_{R[ED_{l}]}}
\end{equation}
where $R[LD]$ and $R[ED_{l}]$ represent the assigned ranks of the $LD$ and $ED_{l}$ vectors, respectively; $\text{cov}(\cdot)$ denotes the covariance of these ranks; and $\sigma$ represents their respective standard deviations.

\begin{figure}
    \centering
    \includegraphics[width=\linewidth]{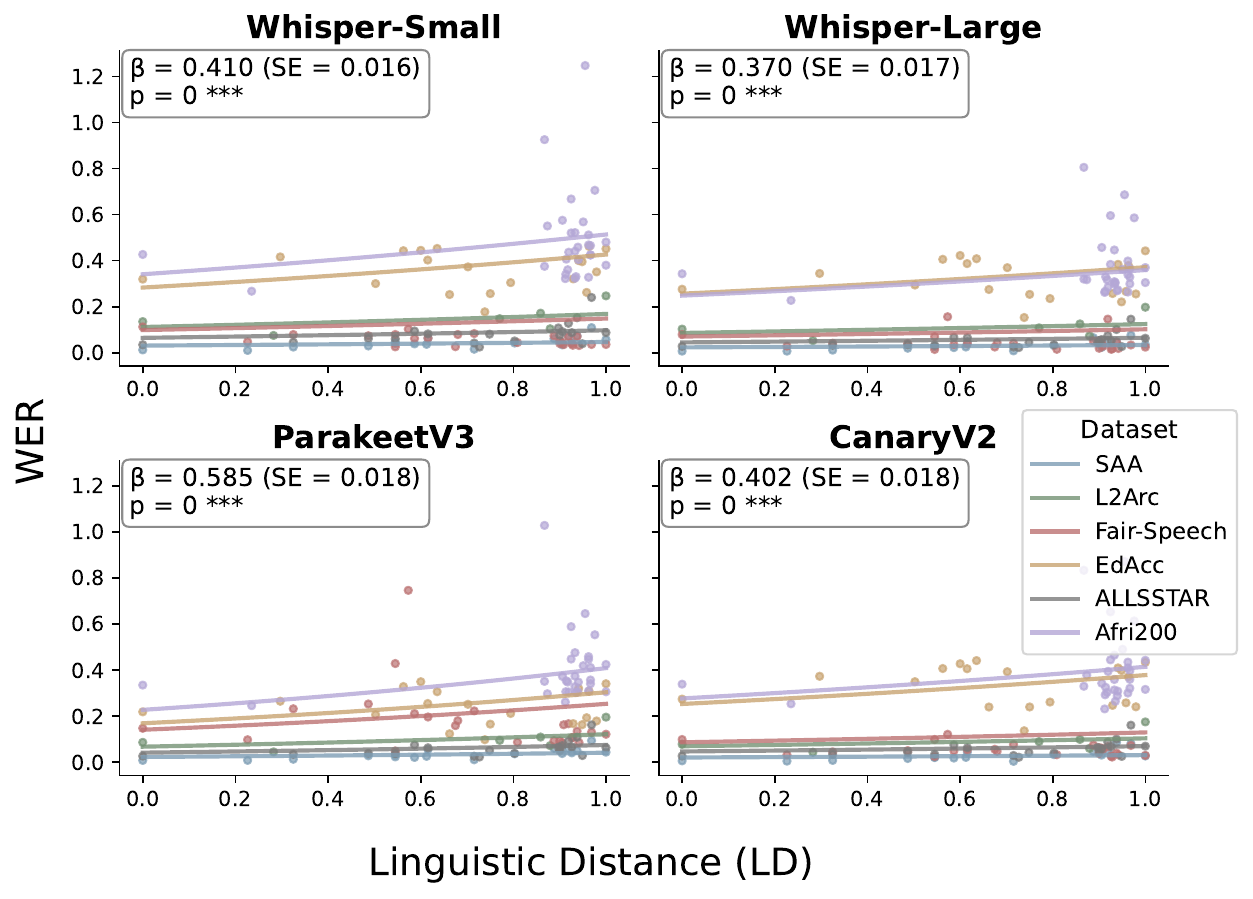}
    \caption{Association between L1's Linguistic Distance ($LD$) from English and ASR performance (WER). Dots represent mean WER for each L1 group, and lines show the fitted relationships between WER and L1's $LD$ from Tweedie mixed-effects models, with dataset as a random effect. All ASR models show a statistically significant positive effect of $LD$ ($p<0.001$). Statistical assumptions were verified and met.}
    \label{fig:MEM}
\end{figure}
\begin{figure*}[htpb]
    \centering
    \includegraphics[width=\linewidth]{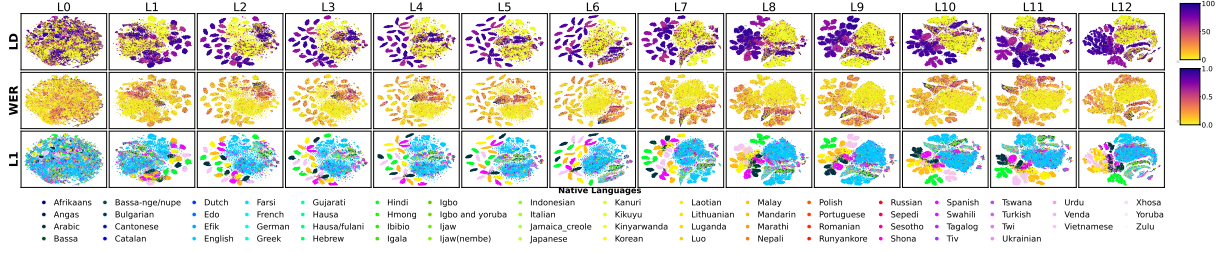}
    \caption{Layer-wise t-SNE visualizations of Whisper Small (perplexity = 30, 1,000 iterations, random state = 20): Linguistic clustering intensifies with model depth, with FairSpeech driving high-WER/low-$LD$ observations.}
    \label{fig:tsne-whisper}
\end{figure*}

\paragraph{Experiment Setup:}
We choose to evaluate English-capable ASR models that are architecturally different to each other, to assess bias against L2 speakers, including the transformer-based Whisper~\cite{radford2022whisper} (Small, 244M, and Large, 1.5B), as well as the Conformer-based Parakeet-TDT-0.6B-v3 and Canary-1B-v2~\cite{sekoyan2025canary}. Table~\ref{tab:dataset_properties} shows the selected datasets during our experiment which are chosen because of their rich L1 backgrounds among speakers who acquired English as a second language.

Beyond the WER, we incorporate Word Information Loss (WIL)~\cite{morris2004and} to better align our results with human perceptual judgment, and Semantic Distance (SemDist)~\cite{kim2021evaluating} to capture errors that may be lexically significant but semantically minor. SemDist, specifically, uses RoBERTa-base to generate semantic embeddings of the reference and hypothesis transcripts, quantifying their similarity via cosine distance.

\section{Results}

\paragraph{Correlation between $LD$ and performance:}
\label{sec:L1L2}

As illustrated in Figure~\ref{fig:coef}, $LD$ exhibits a general positive association with all evaluated error metrics across nearly all datasets, with the sole exception of Fair-Speech. The association is particularly pronounced and statistically significant for the SAA dataset. This pattern suggests that ASR performance tends to deteriorate as the $LD$ between speakers' L1 and English increases. Critically, this pattern remains stable across different model architectures, 
confirming that increased $LD$ from English systematically drives higher downstream recognition error rates. Across all linear regressions, $LD$ accounts for between 1\% and 56\% of the total performance variance. However, $LD$ is an explanatory factor and its explanatory power varies substantially across datasets and model settings: $LD$ explains a larger proportion of the variance in settings with higher adjusted $R^2$ values, while its explanatory power is weaker for datasets such as EdAcc or Afrispeech-200, (see Appendix~\ref{App:LD} for further discussion). Overall, these results suggest that $LD$ exerts a small-to-moderate yet systematic influence on model errors, given that ASR performance is simultaneously shaped by speaker-specific variables such as L2 proficiency, age, and gender.

Conversely, Fair-Speech dataset shows negative regression coefficients across all evaluated models and performance metrics, with the total variance explained by $LD$ ranging between 2\% and 31\%. To investigate this anomaly, we conducted a preliminary human perception study\footnote{Detail of setup is available in Appendix~\ref{App:human}. All relevant statistical assumptions were verified and met.} ($n=4$) to evaluate whether listeners could reliably distinguish between the corpus's designated L1 and L2 speaker cohorts. Our perceptual results reveal that the acoustic boundaries between L1 and L2 speakers in this dataset are remarkably thin. While the average classification accuracy across evaluators was 58.8\%, binomial tests against a chance baseline (50\%) demonstrated that only Evaluator 1 ($p = 0.003$) and Evaluator 3 ($p = 0.016$) performed significantly better than random guessing. Furthermore, the low inter-rater reliability (Fleiss’ $\kappa = 0.29$) confirms a distinct lack of perceptual consensus among the human annotators. However, this should be interpreted as a preliminary perceptual check of whether the dataset's designated L1/L2 distinction was audible to listeners under our experimental task, rather than as an expert validation of speakers' linguistic backgrounds or the accuracy of the dataset labels.

Collectively, these findings suggest that Fair-Speech corpus may contain substantial demographic and sociolinguistic variation that influences the observed relationship between $LD$ and ASR performance. Since all participants, including L2 speakers, are recruited in the United States, their speech may exhibit greater convergence toward general American English. This potentially reduce the retention of the distinct phonological markers associated with their primary language, which may contributing to the perceptual ambiguity documented in our human validation study. At the same time, the substantial diversity within the corpus, including variation among English L1 speakers, suggests that factors beyond L1/L2 status may contribute to the observed WER patterns and may also explain the negative $LD$ coefficients. Therefore, we view Fair-Speech as a counterexample demonstrating that English L1/L2 status and $LD$ are not necessarily the primary axes of ASR disparity in every corpus.

\paragraph{LD-ASR Association Across Datasets:}
To account for systematic differences in ASR performance across datasets arising from differences in recording conditions and task design, we further fitted Tweedie mixed-effects models with dataset-specific random intercepts. 
The results (Fig.~\ref{fig:MEM}) are consistent with our previous analysis, with $LD$ showing a positive and statistically significant association with WER across the evaluated models. For example,
in Whisper-Small, $LD$ has positive association with WER ($\beta=0.410$, $SE = 0.016$, $z=25.20$, $p<0.001$). Given the log link, this corresponds to an approximately 52\% increase in expected WER for a one-unit increase in $LD$, holding the dataset-specific random effect constant. The estimated dataset-level random-effect variance is $\hat{\sigma}_u^2=0.9997$ ($SE=0.702$). Overall, this analysis provides additional support for the association between $LD$ and ASR error, suggesting that the observed relationship persists after accounting dataset-level variation\footnote{The association also holds for WIL and SemDist; see Appendix~\ref{tweedie} for results}.

\paragraph{Model layer depth influence on accent features:}
To investigate the origins of the disparity in $LD$-ASR performance within the latent space, we visualize the embeddings using t-SNE projections to identify potential separation based on speakers' L1. Taking Whisper-small as an example (Figure~\ref{fig:tsne-whisper}), linguistic clustering becomes increasingly pronounced in the deeper layers and remains evident in the final layers, indicating that L1-related structure is preserved in the representations. Furthermore, this clustering within the embedding space emerges as early as the first few layers, aligning with the higher error rates observed in early-stage representations. This phenomenon is not unique to Whisper architecture. A similar layer-wise spatial separation can also be observed in both Parakeet and Canary\footnote{Other models visualizations are provided in Appendix~\ref{App:Emb}}.

\begin{figure}[!htb]
    \centering
    \begin{subfigure}{\linewidth}
        \centering
        \includegraphics[width=\linewidth]{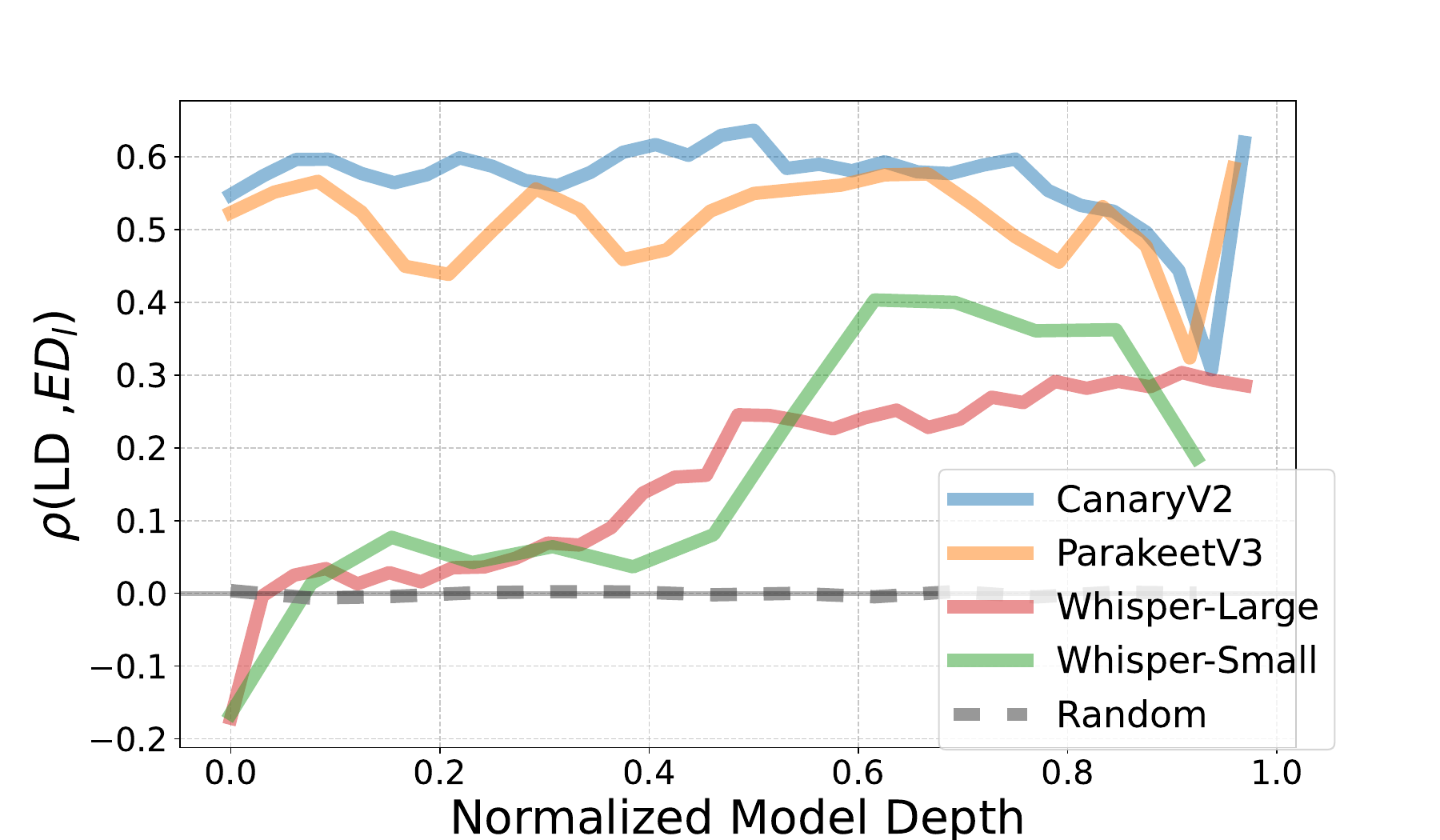}
        \caption{Spearman correlation across Linguistic Distance ($LD$) and Embedding Distance ($ED_{l}$).}
    \end{subfigure}

    \hfill
    \begin{subfigure}{\linewidth}
        \centering
        \includegraphics[width=\linewidth]{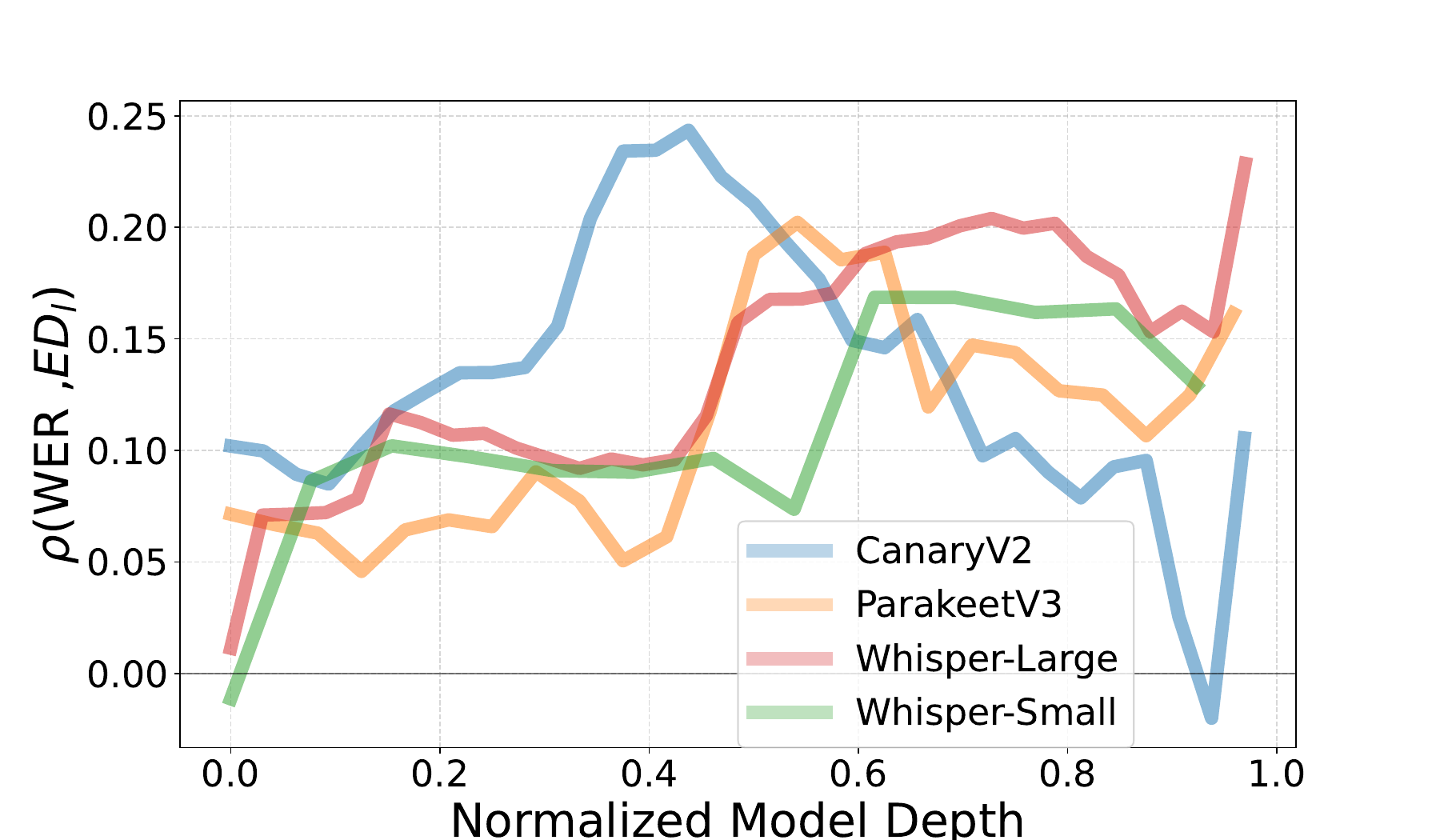}
        \caption{Spearman correlation across Word Error Rates ($Y^\text{WER}$) and Embedding Distance ($ED_{l}$).}
    \end{subfigure}

    \caption{Evolution of Spearman correlation metrics across network layers. Embedding space divergence relative to language distance $\rho(LD, ED_{l})$ and Word Error Rates $\rho(Y^\text{WER}, ED_{l})$ intensify with layer depth. Due to non-normality, Spearman's rank correlation was used; All statistical assumptions were satisfied.}
    \label{fig:corr}
\end{figure}

To examine our hypothesis that $LD$ is encoded across the model layers and is associated with ASR errors, we calculate the Spearmans correlation across both $LD$ and $\text{ED}_{\text{l}}$, and $LD$ and $\text{Y}^\text{metric}$. Figure~\ref{fig:corr} (a) shows that Whisper, no matter the model size, has correlation between $LD$ and $\text{ED}_{\text{l}}$ becomes pronounced from the early-to-middle layers and remains evident in the later layers. This pattern is also reflected in the relationship between $\text{ED}_{\text{l}}$ and $\text{Y}^{\text{WER}}$ showing in Figure~\ref{fig:corr} (b)\footnote{WIL and SemDist results are available in Appendix~\ref{App:cor}}. As for Parakeet/Canary, they both hold a high correlation across $LD$ and $\text{ED}_{\text{l}}$. However, CanaryV2 shows lower correlation across performance metrics. It aligns with Figure~\ref{fig:coef}, where Canary generally has the lowest coefficient comparing to other models. 

\section{Conclusion}

This work explores how Linguistic Distance ($LD$) impacts English Automatic Speech Recognition (ASR) accuracy across different architectures. We reveal that as a speaker's first language (L1) $LD$ from English increase, ASR performance systematically degrades. This relationship remains after accounting for dataset-level variation in mixed-effects models. Fair-Speech provides a counterexample, highlighting that L1/L2 status and $LD$ are not always the primary axes of the disparity. By mapping internal model dynamics, we show that L1/L2-related separation in acoustic embeddings persists across deeper neural network layers, suggesting that models retain rather than normalize accent-related variation. Crucially, CanaryV2 emerges as a notable exception on within this investigation, indicating greater robustness to accent diversity.

\section{Limitations}
While our study incorporates diverse datasets, the L2 samples from available dataset are still primarily drawn from specific geographic populations, particularly European and East Asian speakers. This leaves other regions underrepresented in our current analysis, potentially limiting the generalizability of our findings to all marginalized speaker groups. 

Furthermore, our reliance on $LD$ as a metric may not fully capture the nuance of diverse linguistic backgrounds. Dataset-provided or self-reported L1 labels can be ambiguous and may not provide a complete representation of a speaker's linguistic history. Specifically, the speech patterns of multilingual individuals often reflect blended influences that cannot be easily categorized by a single first language L1, but has been filtered out during the analysis. Treating L1 and $LD$ as primary explanations for ASR disparities may therefore oversimplify complex linguistic histories and overlook substantial within-corpus variability, including L2 proficiency, age of acquisition and exposure to English. These factors may contribute to variation in the observed results and influence the relationship between $LD$ and ASR performance, potentially affecting our conclusions. 

Another limitation of this work is the lack of a universally standardized measure of $LD$. Although the approach proposed by~\citet{vincent2020stochastic} provides a single composite relatedness score that is convenient for our regression analysis, other commonly used resources, such as URIEL/lang2vec and WALS, capture a broader range of linguistic features but require additional choices regarding feature selection and aggregation. Consequently, the observed relationship between $LD$ and ASR performance may depend, to some extent, on the particular distance measure adopted.

In our study, we focus on compare commonly used open-source ASR model. Although the selected models are multilingual or multilingual-capable, they are not designed the same way as massively multilingual system such as Omnilingual, or commericial system. Those models trained with broader multilingual coverage may encode L1-related variation differently and may reduce or reshape the disparities observed in our selected model. 

Finally, since our empirical analysis is fully focuses on English speech, further investigation is still needed before generalize the effect of language distance to other target languages ASR system or broader setting.

\section{Ethics Considerations}
Demonstrating that ASR models separate speakers by L1 in their latent space reveals a dual-use risk. These internal representations could be extracted without consent to profile or surveil users based on their background. Furthermore, we show that optimizing only for output parity (WER, WIL, SemDist) hides this internal bias, allowing structurally biased models to be deployed while falsely appearing fair.



\bibliography{ref}

\appendix
\newpage
\section{Weak Explanatory Power of $LD$}
\label{App:LD}
$LD$ has weaker explanatory power in datasets such as EDACC and Afrispeech-200, where additional sources of variability may influence ASR performance beyond linguistic distance. EDACC contains spontaneous dyadic conversations that includes speaker overlap and background noise, while Afrispeech-200 includes clinical domain speech, for which ASR systems have been shown to perform worse than on more general-domain speech~\cite{olatunji2023afrispeech}. These dataset-specific characteristics reinforce the finding that systematic variation observed in English ASR performance cannot be attributed to $LD$ alone. Instead, the explanatory power of $LD$ should be interpreted alongside dataset design, sociolinguistic and demographic variation, multilingual speaker histories, and other corpus-level factors.

\section{Human Perception Study}
\label{App:human}
To qualitatively verify the Fair-Speech dataset, an informal listening test was conducted with four volunteering participants based in Denmark with an English proficiency of at least C1. Each participant was presented with 100 randomized audio samples, consisting of 50 L1 and 50 L2 speakers. The participants were instructed to label each sample based solely on their perception of the speaker’s language status, where L1 referred to speakers whose first language was English and L2 referred to speakers whose first language was not English. Prior to the task, participants were briefed on the study’s objective: validating ASR model performance across diverse speaker backgrounds. They were also informed that their responses would be fully anonymized and used strictly for internal result verification, with no individual-level data to be publicly released. 
\newpage
\section{Tweedie Mixed Effect Models for Alternative Performance Metrics}
\label{tweedie}
\begin{figure}[H]

    \centering

    \begin{subfigure}{\linewidth}
        \centering
        \includegraphics[width=\linewidth]{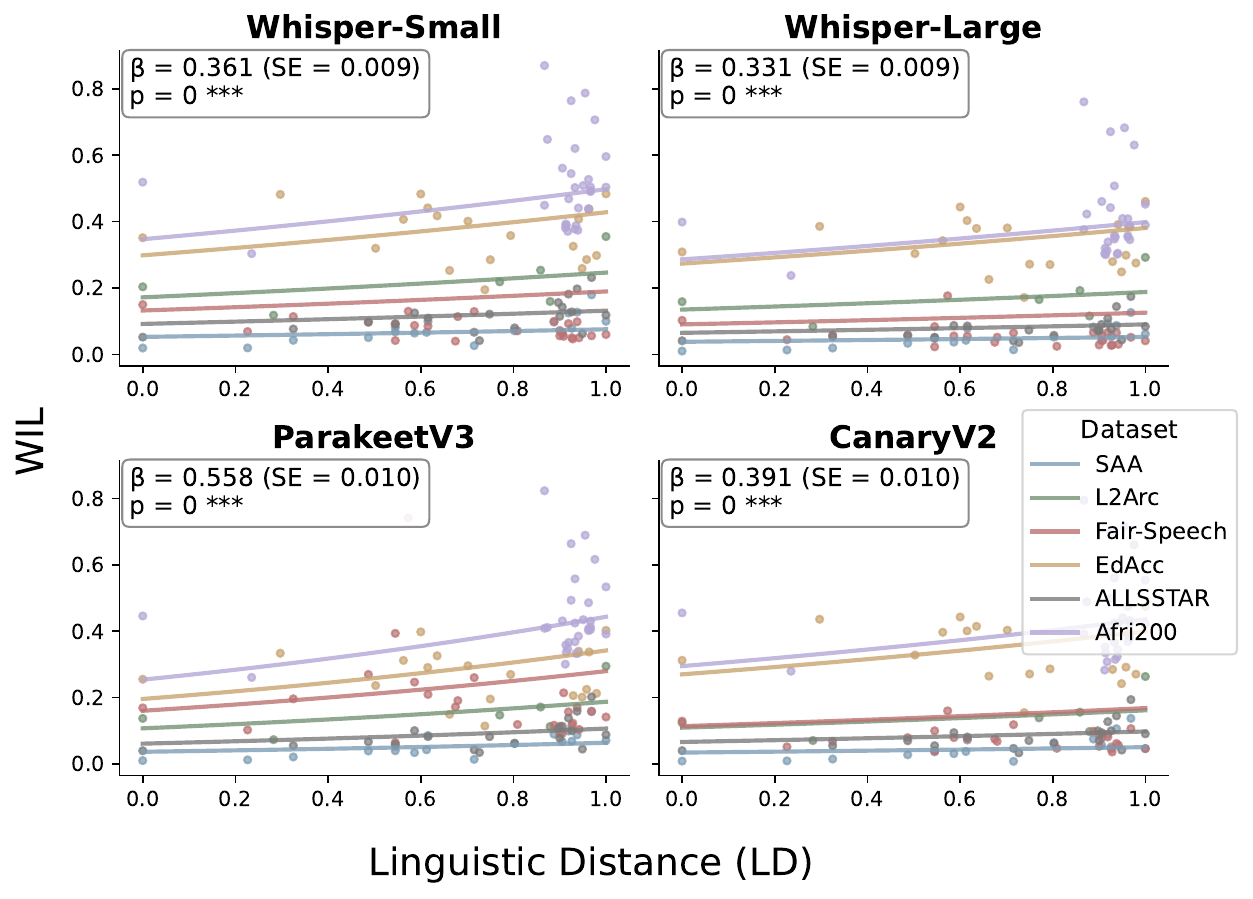}
        \caption{Association between L1’s Linguistic Distance (LD) from English and ASR performance in Word Information Lost ($Y^\text{WIL}$).}
    \end{subfigure}
    \begin{subfigure}{\linewidth}
        \centering
        \includegraphics[width=\linewidth]{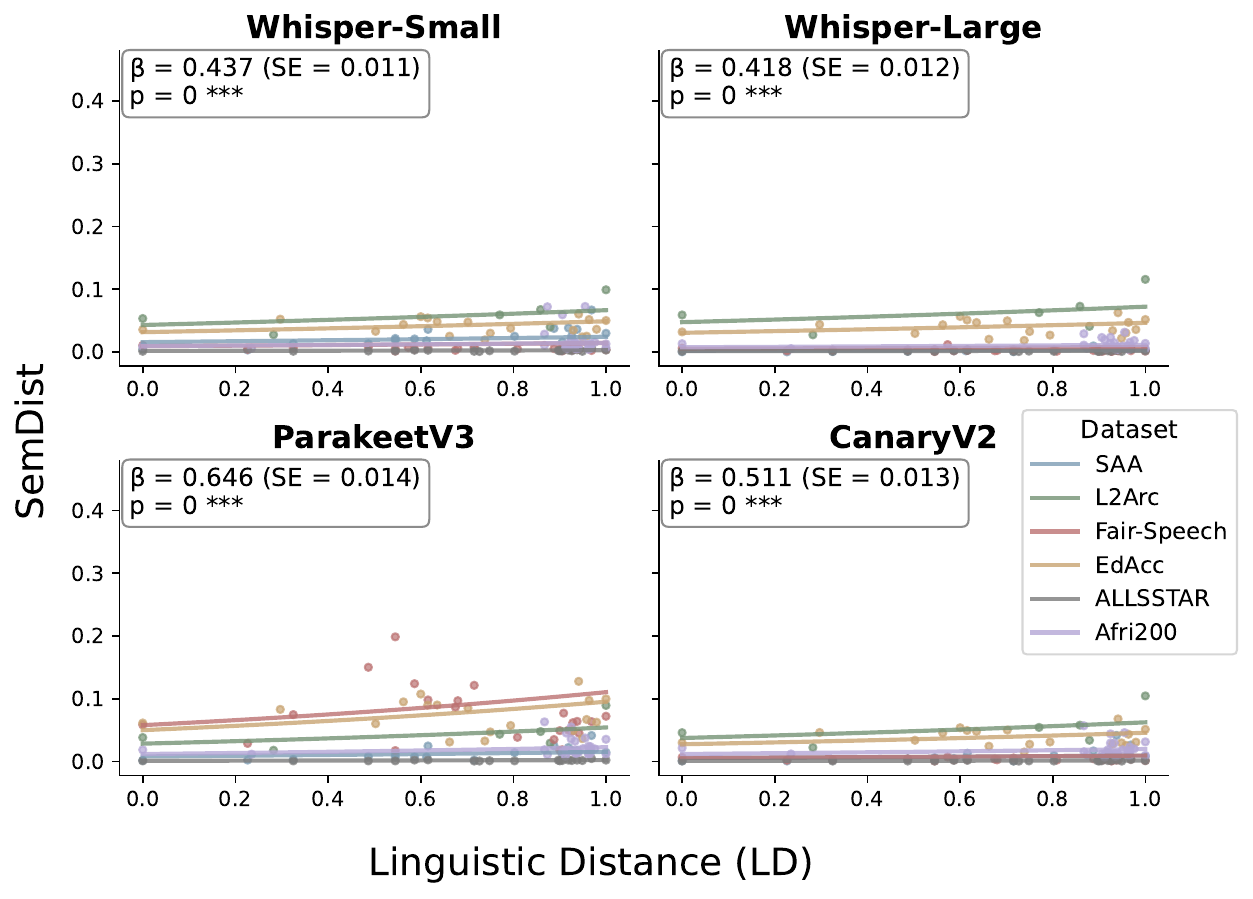}
        \caption{Association between L1’s Linguistic Distance (LD) from English and ASR performance in Semantic Distance (SemDist).}
    \end{subfigure}

    \caption{The association between LD and ASR performance is consistent across alternative evaluation metrics, WIL and SemDist.}
    \label{fig:corr_metric}
\end{figure}
\newpage
\section{Comparative Embedding Analysis Across ASR Architectures}
\label{App:Emb}
\begin{figure}[!htb]
    \centering
    \begin{subfigure}{\linewidth}
        \centering
        \includegraphics[width=\linewidth]{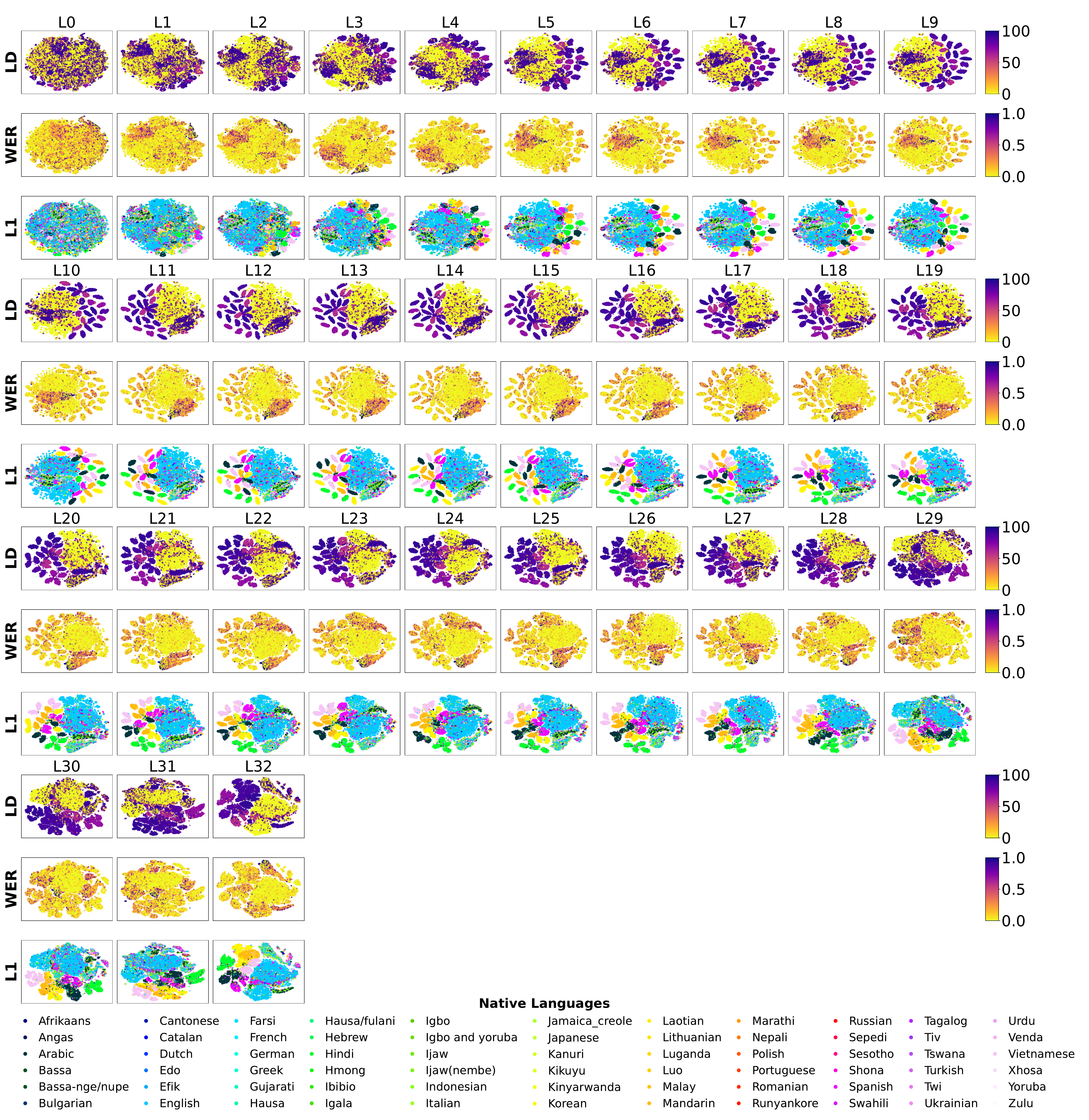}
        \caption{Whisper Large (1.5B)}
    \end{subfigure}
    \hfill
    \begin{subfigure}{\linewidth}
        \centering
        \includegraphics[width=\linewidth]{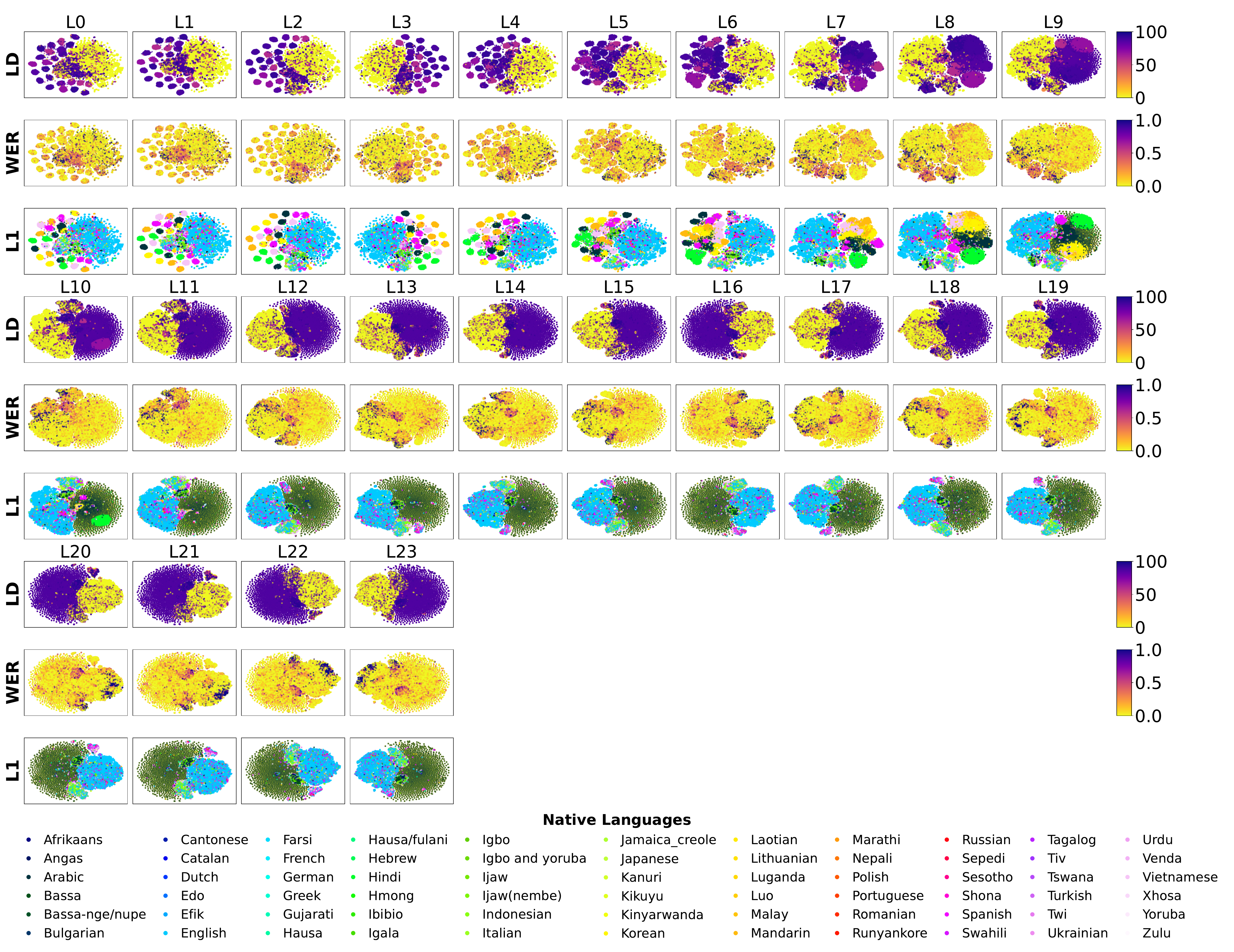}
        \caption{Parakeet-TDT-v3 (0.6B)}
    \end{subfigure}
\end{figure}
\begin{figure}[!htb]
    \centering
    \begin{subfigure}{\linewidth}
        \centering
        \includegraphics[width=\linewidth]{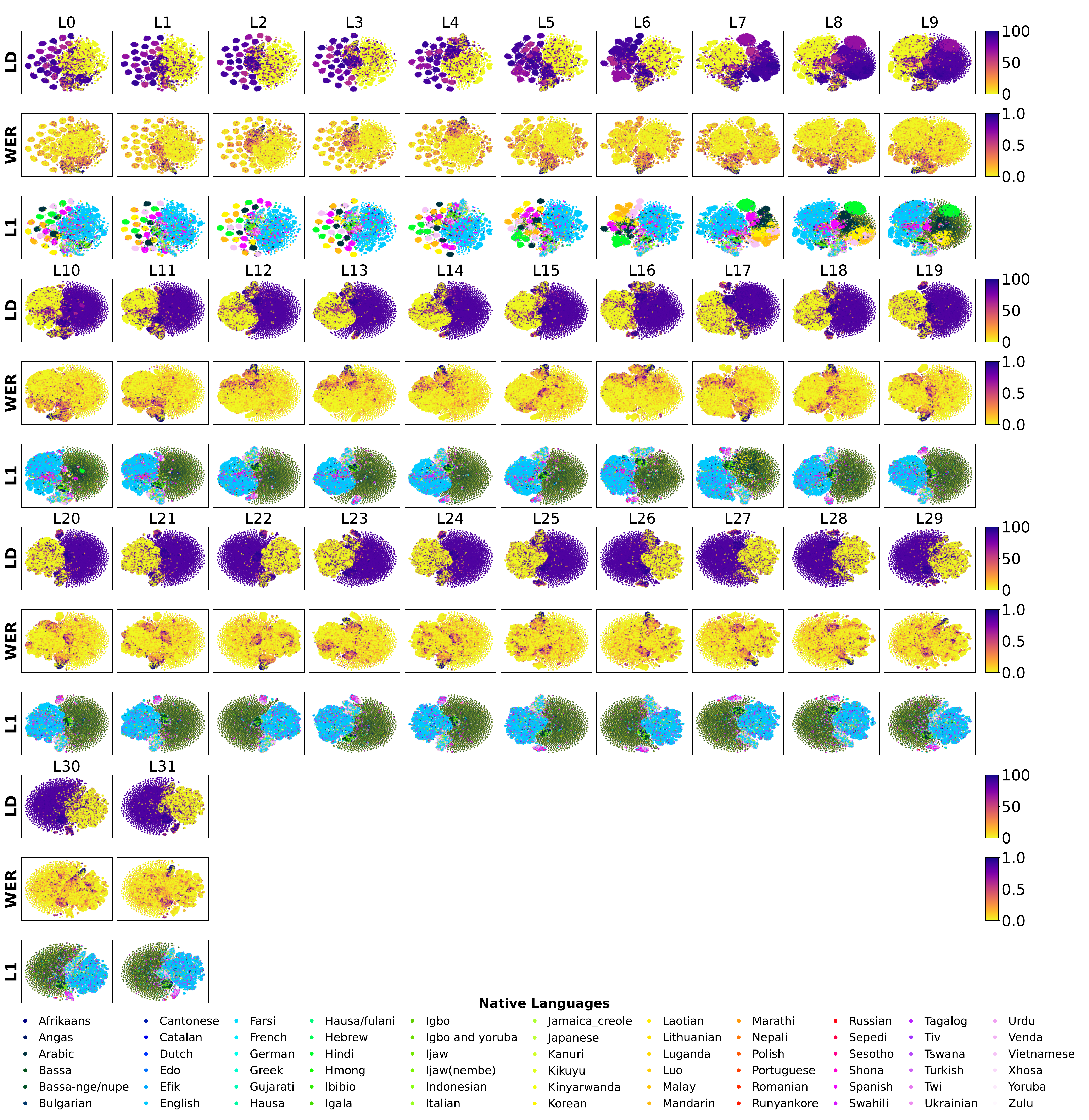}
        \caption{Canary-v2 (1B)}
    \end{subfigure}
    \caption{Layer-wise evolution of latent embeddings. t-SNE (perplexity = 30, 1,000 iterations, random state = 20) projections demonstrate the progressive segregation of L2 speech features across different ASR architectures. The observed patterns are consistent across alternative t-SNE settings (perplexity = 10, random state = 49; perplexity = 30, random state = 42).}
    \label{fig:corr_WIL}
\end{figure}

\newpage
\section{Correlation Analysis for Alternative Performance Metrics}
\label{App:cor}
\begin{figure}[H]

    \centering

    \begin{subfigure}{\linewidth}
        \centering
        \includegraphics[width=\linewidth]{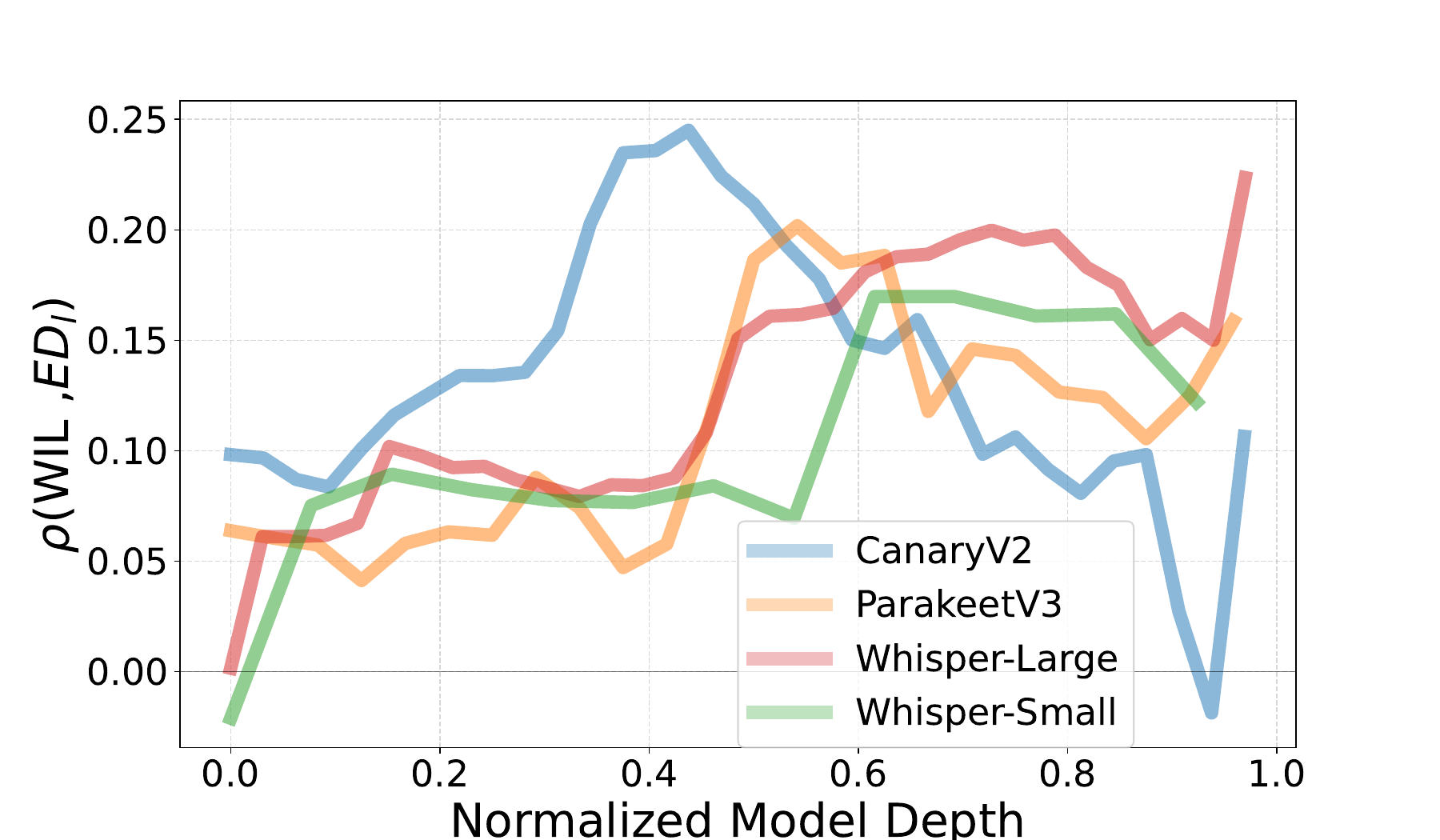}
        \caption{Correlation between Word Information Lost ($Y^\text{WIL}$) and Embedding Distance ($ED_{l}$).}
    \end{subfigure}
    \begin{subfigure}{\linewidth}
        \centering
        \includegraphics[width=\linewidth]{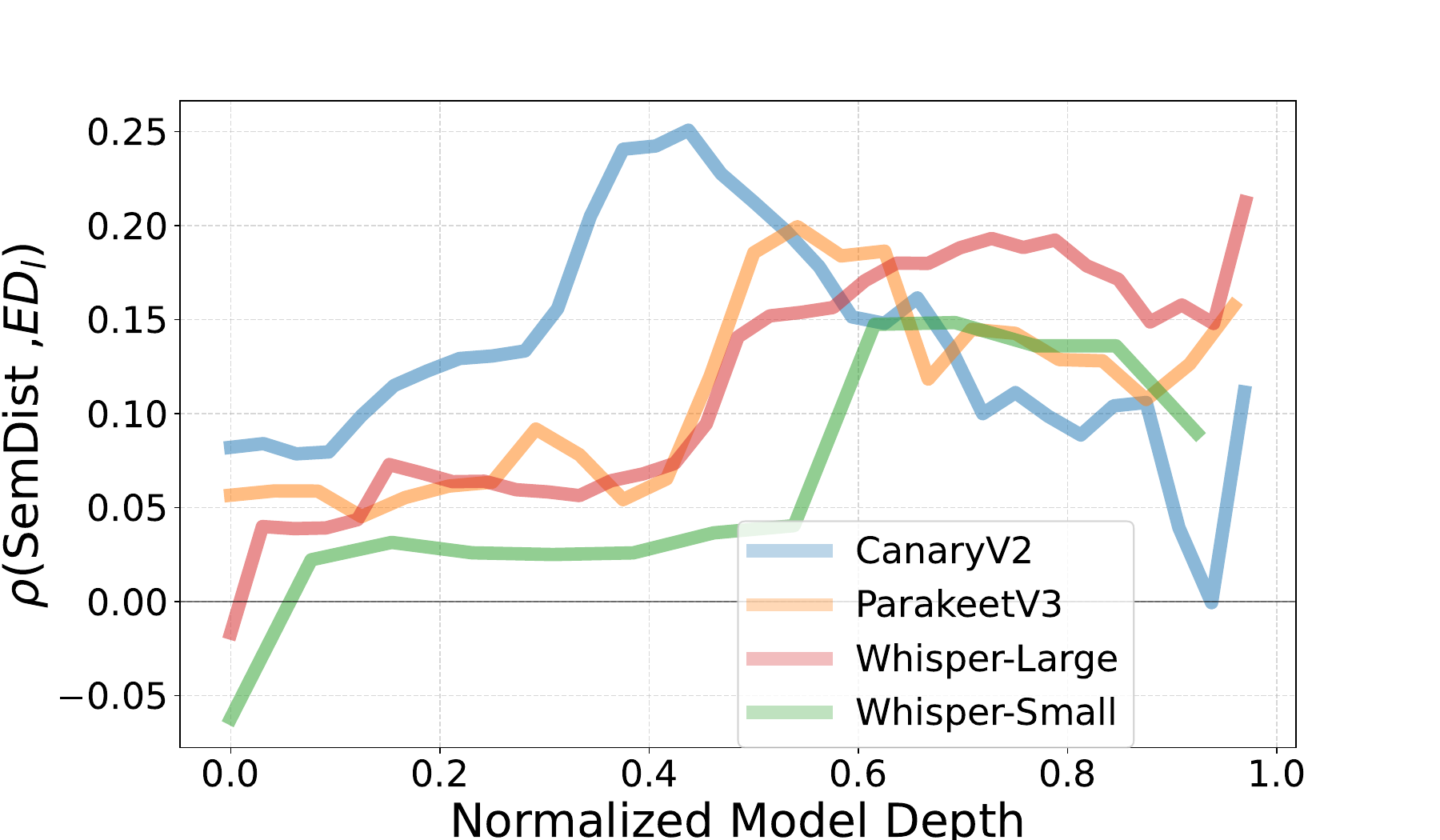}
        \caption{Correlation between Semantic Distance ($Y^\text{SemDist}$) and Embedding Distance ($ED_{l}$).}
    \end{subfigure}

    \caption{Evolution of Spearman correlation metrics across network layers. The plots illustrate consistent alignment between internal representation shifts and alternative performance metrics (WIL and SemDist)}
    \label{fig:corr_metric}
\end{figure}
\newpage

\onecolumn
\section{AI Usage card based on~\cite{wahle2023ai}}

{\sffamily
    \centering
    \tcbset{colback=white!10!white}
    \begin{tcolorbox}[
        title={\large \textbf{AI Usage Card}\hfill \makebox{\qrcode[height=1cm]{https://ai-cards.org/}}},
        breakable,
        boxrule=0.7pt,
        width=.8\paperwidth,
        center,
        skin=bicolor,
        segmentation empty,
        before lower={\footnotesize{AI Usage Card v1.0 \hfill \url{https://ai-cards.org} \hfill \hfill \href{https://ai-cards.org/whitepaper.pdf}{PDF} | \href{https://ai-cards.org/whitepaper.bib}{BibTeX} | \href{https://ai-cards.org/whitepaper.xml}{XML} | \href{https://ai-cards.org/whitepaper.json}{JSON} | \href{https://ai-cards.org/whitepaper.csv}{CSV}}},
        halign lower=center,
        collower=white,
        colbacklower=tcbcolframe]
        
    \footnotesize{
        \begin{longtable}{p{.15\paperwidth} p{.275\paperwidth} p{.275\paperwidth}}
             {\color{LightBlue} \MakeUppercase{Correspondence(s)}} \newline  Ting-Hui Cheng
             & {\color{LightBlue} \MakeUppercase{Contact(s)}} \newline tiche@dtu.dk 
             & {\color{LightBlue} \MakeUppercase{Affiliation(s)}} \newline Technical University of Denmark \\\\
             & {\color{LightBlue} \MakeUppercase{Project Name}} \newline Linguistic Distance Segregates Latent Representations in Automatic Speech Recognition Systems & {\color{LightBlue} \MakeUppercase{Key Application(s)}} \newline Evaluation of ASR robustness \\\\
             {\color{LightBlue} \MakeUppercase{Model(s)}} \newline ChatGPT \newline Gemini \newline Claude
             
             & {\color{LightBlue} \MakeUppercase{Version(s)}} \newline 5.1, 5.6 Luna \newline 2.5 Pro, 3 Pro \newline Sonnet 5 \\\\
             \cmidrule{2-3}\\
             
             {\color{LightBlue} \MakeUppercase{Writing}} \newline ChatGPT, Gemini
             & {\color{LightBlue} \MakeUppercase{Generating new text based on instructions}} \newline  { \color{Grey}Not used} 
             & {\color{LightBlue} \MakeUppercase{Assisting in improving own content}} \newline When existing text is paraphrased or improved. \\\\
             & {\color{LightBlue} \MakeUppercase{Paraphrasing related work}} \newline { \color{Grey}Not used} 

             & {\color{LightBlue} \MakeUppercase{Putting other works in perspective}} \newline { \color{Grey}Not used}   \\\\

             {\color{LightBlue} \MakeUppercase{Coding}} \newline Claude,ChatGPT, Gemini
             & {\color{LightBlue} \MakeUppercase{Generating new code based on descriptions or existing code}} \newline When new code is generated based on instructions or prompts. 
             & {\color{LightBlue} \MakeUppercase{Refactoring and optimizing existing code}} \newline When existing or generated code is refactored or its performance optimized. \\\\
             & {\color{LightBlue} \MakeUppercase{Comparing aspects of existing code}} \newline  { \color{Grey}Not used}  \\\\
             
\cmidrule{2-3}\\
             
             {\color{LightBlue} \MakeUppercase{Ethics}}
             & {\color{LightBlue} \MakeUppercase{What are the implications of using AI for this project?}} \newline To facilitate academic writing and improve efficiency.
             & {\color{LightBlue} \MakeUppercase{What steps are we taking to mitigate errors of AI for this project?}} \newline All AI-assisted content is critically reviewed and verified by the authors. \\\\
             & {\color{LightBlue} \MakeUppercase{What steps are we taking to minimize the chance of harm or inappropriate use of AI for this project?}} \newline Only to support writing and coding assistance and not used to generate experimental results or scientific conclusions.
             & {\color{LightBlue} \MakeUppercase{The corresponding authors verify and agree with the modifications or generations of their  used AI-generated content}} \newline Yes   \\
        \end{longtable}
     }
    \tcblower
    \end{tcolorbox}
}

    

\end{document}